\documentclass[runningheads]{llncs}
\usepackage{graphicx}

\usepackage{tikz}
\usepackage{booktabs}
\usepackage{hyperref}
\usepackage{comment}
\usepackage{amsmath,amssymb} 
\usepackage{subfiles}

\usepackage[accsupp]{axessibility}  

\makeatletter
\def\blfootnote{\gdef\@thefnmark{}\@footnotetext}

\newcommand{\superscript}[1]{\ensuremath{^{\textrm{#1}}}}

\begin{document}
\pagestyle{headings}
\mainmatter
\title{Towards Learning Neural Representations from Shadows} 


\titlerunning{Towards Learning Neural Representations from Shadows}

\author{Kushagra Tiwary \superscript{*} \and
Tzofi Klinghoffer \superscript{*} \and
Ramesh Raskar}

\authorrunning{Tiwary K. et al.}
%
\institute{Massachusetts Institute of Technology \\
\email{\{ktiwary, tzofi, raskar\}@mit.edu}
}
\maketitle


\begin{abstract}
We present a method that learns neural shadow fields which are neural scene representations that are \emph{only} learnt from the shadows present in the scene. While traditional shape-from-shadow (SfS) algorithms reconstruct geometry from shadows, they assume a fixed scanning setup and fail to generalize to complex scenes. Neural rendering algorithms, on the other hand, rely on photometric consistency between RGB images, but largely ignore physical cues such as shadows, which have been shown to provide valuable information about the scene. We observe that shadows are a powerful cue that can constrain neural scene representations to \emph{learn} SfS, and even outperform NeRF to reconstruct otherwise hidden geometry. We propose a graphics-inspired differentiable approach to render accurate shadows with volumetric rendering, predicting a shadow map that can be compared to the ground truth shadow. Even with just binary shadow maps, we show that neural rendering can localize the object and estimate coarse geometry. Our approach reveals that sparse cues in images can be used to estimate geometry using differentiable volumetric rendering. Moreover, our framework is highly generalizable and can work alongside existing 3D reconstruction techniques that otherwise only use photometric consistency. 

\keywords{Scene Representations, Differentiable Rendering, 3D Scene Reconstruction, Shape-from-Shadows, Volume Rendering}
\end{abstract}

\section{Introduction}
Recovering 3D geometry from 2D images remains an extremely important, yet unsolved problem in computer vision and inverse graphics. Considerable progress has been made in the field when assumptions are made, such as bounded scenes, diffuse surfaces, and specific materials. However, reconstruction algorithms still remain largely susceptible to real world effects, such as specularity, shadows, and occlusions \cite{sfs_real_world_1}. This susceptibility is largely due the variation in different materials and textures, and a non-unique mapping from 3D geometries to 2D images. Even though these effects cause issues for many methods, they also provide valuable information about the scene and geometry of the object. For example, cues like self-shadows provide vital information about an object's concavities, while shadows cast on the ground plane provide information about its geometry. Moreover, shadows are independent of textures and surface reflectance models and are a strong cue in overhead imagery where vertical surfaces, like facades, are sampled poorly, whereas oblique lighting can expose this geometry. Exploiting, instead of ignoring these cues, can make algorithms robust and the fundamental problem of 3D reconstruction less ill-posed. 

Previous works in recovering 3D shape of objects by exploiting physical cues has relied on constructing inverse models to explicitly handle and exploit cues such as shadows, shading, motion, or polarization \cite{6278478} \cite{85658} \cite{784284}. These approaches are physically anchored as they use properties of light or surface reflectance models to exploit cues and only need up to a single image to reconstruct simple objects. Albeit successful under strict assumptions about lighting, camera, and the object, these models typically cannot handle complex scenes and do not translate well into real-world scenarios as creating inverse models to capture complex physical phenomenon soon becomes intractable and hard to optimize.

\blfootnote{\superscript{$*$} Equal contribution}

\begin{figure}[t]
\begin{center}
   \includegraphics[width=1\linewidth]{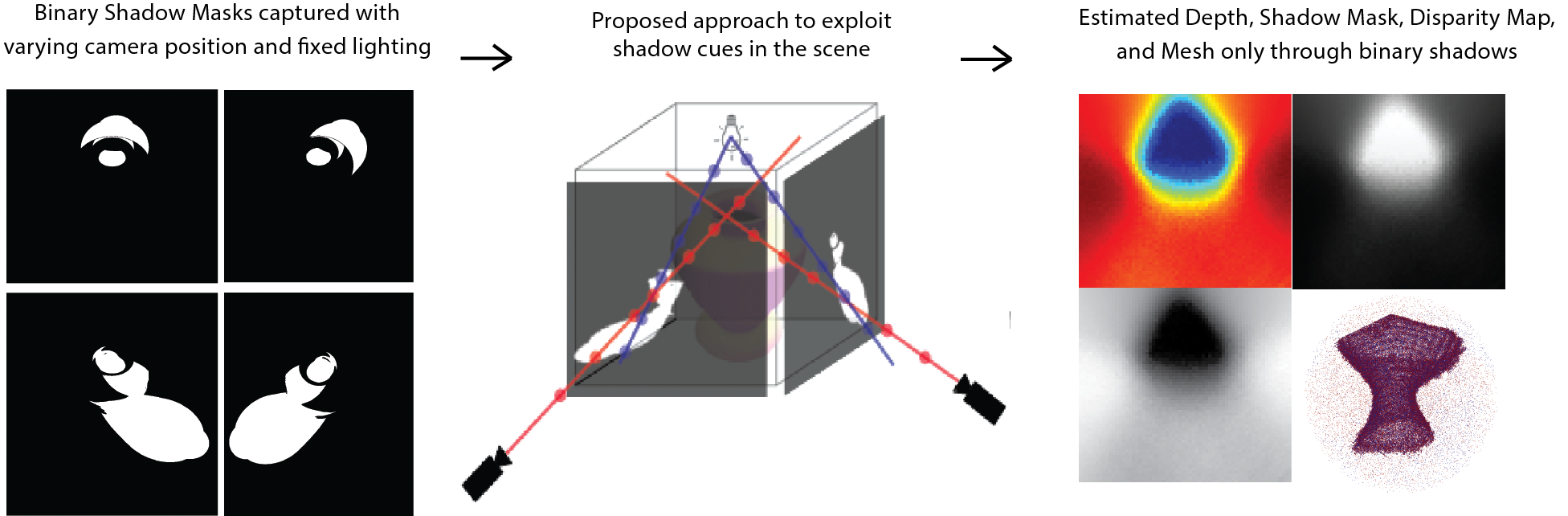}
\end{center}
\caption{\textbf{Exploiting physical cues in neural rendering.} Our approach takes sparse binary shadow masks captured with varying camera positions under fixed lighting and uses our proposed differentiable shadow rendering model to estimate shadow maps, thereby learning neural scene representations. We can visualize the learned implicit representations by rendering estimated depth maps and estimated shadow maps from novel views. We also run marching cubes \cite{lorensen1987marching} on our learned representations to get explicit meshes for a quantitative analysis.}
\label{fig:teaser_figure}
\end{figure}
To combat the problem of real world variability, modern methods such as \cite{mvcTulsiani18} \cite{nerf} \cite{nerf5} \cite{sitzmann2019deepvoxels} \cite{ye2021shelf} \cite{Lombardi:2019} have largely been data-driven by directly learning 3D representations on real-world scenes based on photometric consistency. Such methods employ an \textit{analysis-by-synthesis} approach to solve the problem by using machine learning to search the space of possible 3D geometries and an inverse model to synthesize the scene based on the predicted geometries. These approaches typically only optimize the photometric loss between different camera viewpoints and show success in learning implicit representation by rendering novel views. However, because they do not explicitly handle these physical cues in their forward model, they fail in scenarios with complex lighting \cite{nerv}, specularity \cite{zhang2021ners}, or reflections \cite{DBLP}.

Motivated by the above observations, we explore what can be learned by exploiting physical cues in a data-driven neural rendering framework. In this paper, we investigate whether the neural rendering framework can learn geometry from physical cues without the assumptions made by the aforementioned methods. We study the use of shadows cast by objects onto themselves and nearby surfaces as the only source of information for 3D reconstruction. While modern approaches for 3D reconstruction ignore such cues, we aim to exploit them. Our unsupervised approach uses \emph{only} shadows to reconstruct the scene by leveraging recent advances in volumetric rendering and machine learning, and therefore proposes a physically anchored data-driven framework to the problem of shape from shadows. Moreover, unlike previous work in shape from shadows, we present a novel method that uses differentiable rendering in the loop to iteratively reconstruct the object based on a loss function instead of iteratively refining the object through explicit carving. Specifically, we use an efficient shadow rendering technique called shadow mapping as the forward model and make it differentiable so that it can be used as an inverse model to iteratively reconstruct the object. Our work also reveals that from limited cues the differentiable volumetric rendering component can \textit{quickly converge to localize and reconstruct a coarse estimate of the object when such cues are explicitly modeled by a forward model}. Our work also suggests that neural rendering can exploit shadows to recover hidden geometry, which otherwise may not be discovered by photometric cues. 

\subsection{Contributions}
\label{sec:Contributions}

\begin{figure}[t]
\centering
\includegraphics[height=4.5cm]{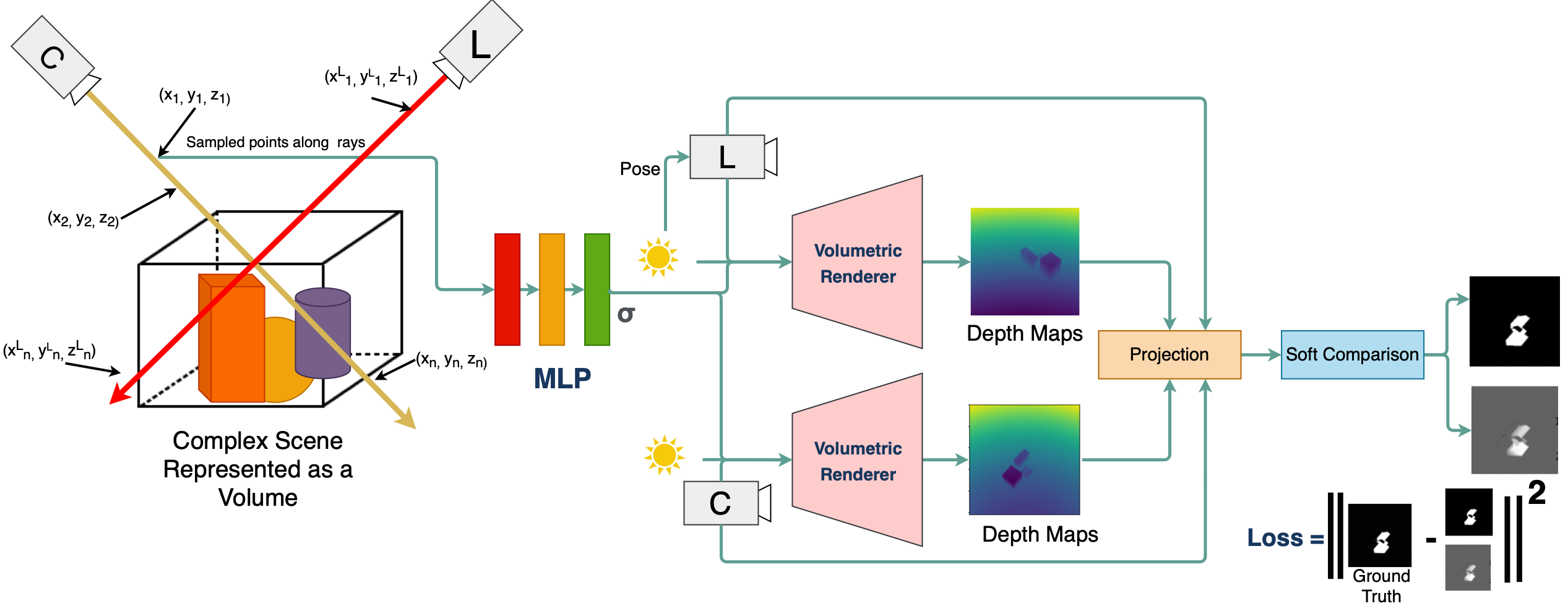}
\caption{\textbf{Overview of the proposed pipeline} We train a neural network to predict opacity at points along the camera and light rays. The opacities are used by the volumetric renderer to output the ray-termination distance which we use to estimate the \emph{z-buffer} from the camera and the light perspective, the latter also known as the shadow map. The estimated \emph{z-buffer} is fed into a \textbf{Projection} step that projects the camera pixels and their associated depths into the light's reference frame. The shadow map is indexed to obtain the corresponding depth values at these new points. The projected depths and indexed depths go through a \textbf{Soft Comparison} step which outputs predicted cast shadows in the scene from the camera's perspective. A loss is computed on the \emph{predicted} and the \emph{ground-truth} shadow mask.}
\label{fig:vol_rend_pipeline}
\end{figure}

Our contributions in this paper are the following: 
\begin{itemize}
\item A framework that directly exploits physical cues like shadows in neural renderers to recover scene geometry. 
\item A novel technique that integrates volumetric rendering with a graphics-inspired forward model to render shadows in an end-to-end differentiable manner. 
\item Results showing that our framework can learn coarse scene representations from just shadows masks. We evaluate the learned representations qualitatively and quantitatively against vanilla neural rendering approaches. To the best of our knowledge, we are the first to show that it is possible to learn neural scene representations from binary shadow masks.
\end{itemize}

\section{Related Work}
\label{sec:Related Work}
\textbf{Shape from Shadows.} Shadowgram imaging deals with estimating the shape of an object through a sequence of shadow masks captured with light sources at various locations. These methods typically assume a controlled and fixed object scanning setup ~\cite{savarese2001shadow} \cite{article}. Martin \& Aggarwal \cite{4767367} introduced a volumetric space carving approach to SfS which outputs a visual hull around the object by carving out voxels lying outside the visual cone. Other work takes a more probabilistic approach to the shape-from-silhouettes problem to make the algorithm more robust to errors \cite{Landabaso2008ShapeFI}. However, interpreting shadows as silhouettes means that self-shadows are not handled, thus motivating Savarese et al. \cite{savarese2001shadow} to propose a method to ``carve" out objects based on self-shadows to create more complete reconstructions.

In contrast, our work takes a differentiable approach to solving the problem through learning. Instead of an explicit carving of voxels we first construct a differentiable forward model that casts shadows based on some geometry. Then, we let the machine learning component predict geometry, which is synthesized by the renderer to cast shadows. Finally, we optimize this setup based on a mean square error between predicted and ground truth shadow masks.  
\begin{figure}[t]
\begin{center}
   \includegraphics[width=0.7\linewidth]{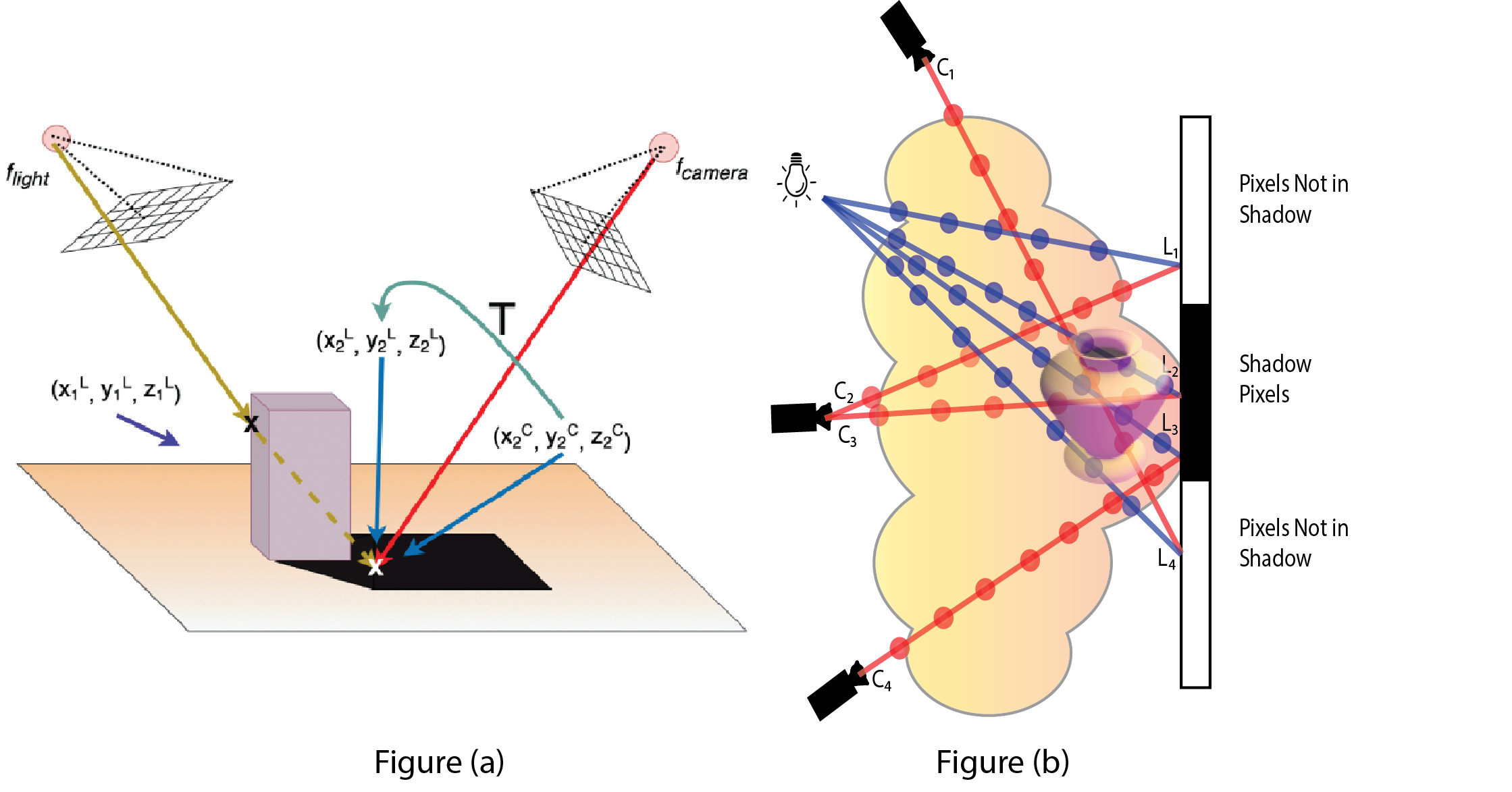}
\end{center}
\caption{Figure \textbf{(a)}: A point \textbf{x} $\in \mathbb{R}^3$ in the scene is defined to be in shadow if no direct path exists from the point \textbf{x} to the light source, implying that there \textbf{must} be an occluding surface between \textbf{x} and the light source. We differentiably render the scene's depth from the camera and the light's perspective at each pixel and then project the camera pixel and its depth into the light's frame of reference. We then  index the light's depth map, or z-buffer, to get $z_1^L$. We note that $z_1^L$ is less than $z_2^L$, i.e. there must be an occluding surface as a ray projected from the light's perspective terminates early. This implies that this point is in shadow. Figure \textbf{(b)} shows a 2D slice of our approach and represents a volume (cloud) with the shadow mask unraveled. The network learns an opacity per point (dots) via the shadow mapping objective which penalizes predicted geometries that don't cast perfect ground truth shadows. Through this, the networks learns 3D geometry that is consistent across all shadows maps for all cameras given a particular light source.}
\label{fig:shadow_mapping}
\end{figure}

\noindent
\textbf{Neural Rendering} Broadly speaking, a neural rendering framework is composed of a differentiable renderer, which can render the scene based on input parameters and is able to differentiate the scene w.r.t. those input parameters. While there are many formulations of differentiable renderers \cite{nimier2019mitsuba} \cite{opendr} \cite{redner} \cite{NMR} that can synthesize  scenes, state-of-art approaches have shown tremendous success by relying on differentiable volumetric rendering \cite{DVR}. Volumetric rendering approaches can realistically render complex scenes and are gradient-friendly. Thus, typical approaches train a neural network to encode the scene and optimize it for photometric consistency between input 2D images from different viewpoints \cite{nerf} \cite{sitzmann2019deepvoxels} \cite{nerf2} \cite{nerf3}. Recent methods such as \cite{DBLP} \cite{verbin2021refnerf} \cite{nerv} \cite{boss2021nerd} explicitly account for specularity, reflections and other such phenomenon, however, the goal of these works are to improve novel view synthesis. Thus, these methods still rely on learning the scene using photometric information.

In contrast, our work deals with 3D reconstruction, not novel view synthesis, and explores what can be learned by relying on shadow cues in the scene. Our framework only operates in the shadow input and output space to infer a 3D representation of the scene. In addition, similar to \cite{yu_and_fridovichkeil2021plenoxels} \cite{yu2021plenoctrees}, our work also reveals that differentiable volumetric rendering is a powerful component that can learn the scene by only relying on sparse physical cues. While volumetric approaches rely on a photometric cues, differentiable rasterization \cite{NMR} \cite{liu2019soft} has been shown to reconstruct 3D mesh using single low dimensional images of ShapeNet objects \cite{shapenet2015} by only using silhouettes. However, these methods fail to show success on high dimensional images, while our approach can scale up to higher dimensional images. Concurrent work by Liu \textit{et al.} \cite{liu2022shadows} also leverages shadows to perform 3D reconstruction, but integrates learned object priors, whereas we solely rely on binary shadow masks and use volumetric rendering.

\noindent
\textbf{Shadows in Graphics.} Graphics deals with the forward model and shadow mapping ~\cite{williams1978casting} is one of the most efficient techniques to render shadows in a scene given the scene's geometry, camera viewpoint and light position. While differentiability is not important for graphics, we make the shadow mapping framework differentiable to work with modern 3D reconstruction algorithms. We describe the algorithm and our implementation in Section \ref{sec:Method}. 

\section{Neural Representations From Shadows}
\label{sec:Method}
Our goal is to recover the scene through shadows cast on the other objects or onto itself. Our method recovers shadows in an image by applying a threshold on that image thereby making no distinction between types of shadows. We show how we model the shape-from-shadows problem using differentiable rendering and implicit representations in Section \ref{sec:shapes_as_nerf} and our graphics-inspired differentiable forward model in Section \ref{sec:diff_sm}. In Section \ref{sec::optimization}, we discuss our additional techniques that we use to enable optimization on binary shadow masks. 

\subsection{Scenes as Neural Shadow Fields}
\label{sec:shapes_as_nerf}

\textbf{Implicit Scene Representations.} Similar to Mildenhall \emph{et al.} \cite{nerf}, we represent a continuous scene by parametrizing it using a learnable function $f_{\theta}$. However, our approach does not include any photometric component, therefore we represent the scene as a 3D function with input $\textbf{x}= (x,y,z)$ and a volumetric density $\sigma$ as output. 

\begin{equation}
\begin{gathered} 
\label{eq:ftheta}
\gamma(\textbf{x}) = \bigg(\sin(2^0\pi \textbf{x}), \cos(2^1\pi \textbf{x}), ..., \sin(2^L-1\pi \textbf{x}), \cos(2^L-1\pi \textbf{x})\bigg) \\
f_{\theta}: \mathbb{R}^{L} \xrightarrow[]{} \mathbb{R}^{+}; (\gamma(\textbf{x})) \mapsto (\sigma)
\end{gathered} 
\end{equation}

\noindent
We use a positional-encoded 3D point $\gamma(\textbf{x}), \{ \gamma(\textbf{x}) \in \mathbb{R}^L, \textbf{x} \in \mathbb{R}^3\}$ as input, which maps to an associated volumetric density $\sigma \in \mathbb{R}^+$ \cite{nerf} \cite{tancik2020fourier}. In contrast, $f$ does not encode view dependant color and is independent to viewing direction. 

\noindent
\textbf{Volumetric Renderer.} We define a volumetric renderer $\mathbf{R}_{\text{vol}}$ that takes $N$ opacities $\{\sigma\}_{i=1}^{N}$ at $N$ discretely sampled points $\{\textbf{x}\}_{i=1}^{N}$ along a ray $\textbf{r}$.
\begin{equation}
\begin{gathered} 
\label{eq::vol}
\mathbf{R}_{\text{vol}}: \big[\mathbb{R}^{+}\big]_{i=1}^N \xrightarrow[]{} \big[\mathbb{R}^+\big]_{i=1}^N; (\{\sigma\}_{i=1}^{N}) \mapsto (\textbf{d})
\end{gathered} 
\end{equation}
\noindent
Since we only have binary shadows as input, we modify the renderer to output the ray termination distance, $\textbf{d}$, instead of the radiance at that ray. $\mathbf{R}_{\text{vol}}$ is not a trainable component, but the ray termination distance, $\textbf{d}$, is differentiable w.r.t. the input opacities. The estimated ray-termination distance, range, is computed as follows:
\begin{equation}
\begin{gathered} 
\label{eq::depth_eq}
\hat{\textbf{D}}(\textbf{r}) = \sum_{i=1}^N T_i\alpha_i t_i; T_i = \prod_{j=1}^{i-1} \big ( 1- \alpha_j \big); \alpha_i = \big(1- e^{-\sigma_i \delta_i} \big)
\end{gathered} 
\end{equation}
\noindent
We sample $\textbf{r}(t)$ at points $\{ t_0, ..., t_N \}$ and evaluate the function ${\textbf{r}}(t) = \textbf{o} + t\textbf{d}$ to get sampled points $\{ x_0, ..., x_N \}$ in the scene. $T_i$ is defined as the cumulative transmittance from $t_0$ until $t_i$ and $\delta_i = t_{i+1}-t_i$ which is the distance between two samples. $\sigma_i$ is the estimated opacity at point $i$ by a learned function $f_{\theta}$. Intuitively, the renderer gives us the ray termination distance for each ray shooting through a pixel.

\begin{figure}[t]
\centering
\includegraphics[width=0.6\linewidth]{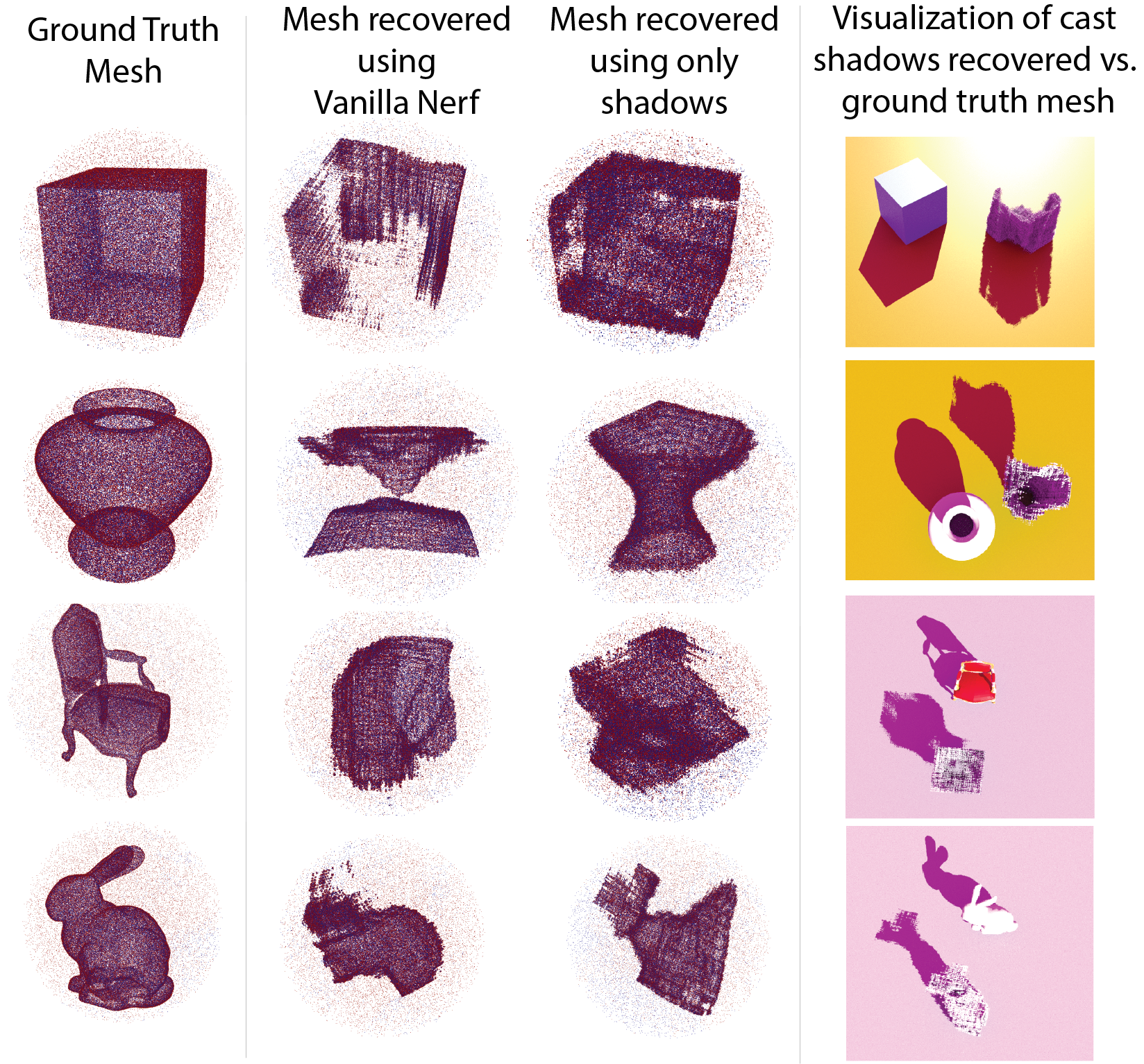}
\caption{\textbf{Qualitative Results.} We observe that for overhead views of the scene where the vertical surface of the vase is sampled poorly in the RGB space, vanilla NeRF fails to exploit geometry cues hidden in cast shadows compared to our approach. Our method doesn't impose any object priors therefore it infers a geometry that will minimize the difference between the predicted and true shadow. Column 4 illustrates that rendered shadows are very similar, indicating that the differentiable rendering framework can indeed learn geometry from sparse shadow cues. Some parts of the objects such as the upper face of cuboid are never in shadow, therefore our approach yields no reconstruction for those surfaces, further showing that the geometry is indeed \emph{only} learnt from cast shadows. We extract the mesh from the volume using marching cubes and visualize it here using a point-cloud SDF representation.}
\label{fig::results_1}
\end{figure}

\subsection{Differentiable Shadow Mapping}
\label{sec:diff_sm}

We define any point \textbf{x} $\in \mathbb{R}^3$ in the scene to be in shadow if no direct path exists from point \textbf{x} to the light source $\mathbf{L}$. This logic implies that there \textbf{must} be some object or an occluding surface between the point \textbf{x} and $\mathbf{L}$ that occludes the light ray from reaching point \textbf{x}. In graphics, shadow mapping \cite{williams1978casting} uses this observation to construct a forward model to render efficient and accurate shadows in the scene based on known light and camera sources. Our approach makes this efficient shadow rendering forward model differentiable so that it can be used as an inverse model. We then pose the problem of shape from shadows and use our proposed inverse model to estimate the 3D geometry of the scene. 

\noindent
\textbf{Estimated z-buffer.} We first evaluate the renderer from the camera's perspective to get the estimated ray termination distance, or range map, $\hat{\textbf{D}}_{cam}$ for all rays coming out of the binary shadow map. However, shadow mapping requires the depth perpendicular to the image plane, i.e along the z axis of the camera’s local coordinate system. This depth is equivalent to a \emph{z-buffer} in graphics and we refer to this value as the \emph{depth} at that pixel. We define a function $g$ to estimate the \emph{z-buffer} $\hat{\textbf{Z}}$ from the range map $\hat{\textbf{D}}$. 
\begin{equation}
\begin{gathered} 
\label{eq::depth_eq_2}
\hat{\textbf{z}}_{u,v} = g(\textbf{d}_{u,v}) = \frac{\textbf{d}_{u,v}}{||(u,v,1) \cdot \mathcal{E}||_2}
\end{gathered} 
\end{equation}
\noindent
The function takes a ray shooting from a pixel $(u,v)$ and a predicted range, $\hat{\textbf{D}}_{cam}^{u,v}$ as input. $\mathcal{E}$ is the rotational component of the camera's extrinsic matrix, $\textbf{d}_{u,v}$ is the ray termination distance from camera's focal point, and $\hat{\textbf{z}}_{u,v}$ is the depth along the z-axis from the pixel $(u,v)$. We also compute the estimated z-buffer from the light's perspective, which we refer to as the estimated \textit{shadow map}. 

\noindent
\textbf{Projection.} With the estimated depths at each pixel from the camera and the light source, we now need to estimate which camera pixels are in shadow given the particular light source. As illustrated in Figure \ref{fig:shadow_mapping}, we do this by projecting all pixels and their associated depths visible by the camera into the light's frame of reference. We then use this projected coordinate to index the shadow map to get the depth to that point from the light's perspective. We formally write this as follows:
\begin{equation}
\begin{gathered} 
\label{eq::project}
(U^l_{cam}, V^l_{cam}, \hat{\textbf{Z}^l}_{cam}) = (U_{cam}, V_{cam}, \hat{\textbf{Z}}_{cam}) \cdot P_{light\_from\_cam} \\ 
\hat{\textbf{Z}}^{U^l_{c}, V^l_{c}}_{light} = \hat{\textbf{Z}}_{light}\bigg[U^l_{cam}, V^l_{cam}\bigg]
\end{gathered} 
\end{equation}
Here, $\hat{\textbf{Z}}_{cam} \in \mathbb{R}^{H\times W}$ is the estimated z-buffer from the camera's perspective at pixels $\{U_{cam}, V_{cam}\}\in \mathbb{R}^{H\times W}$. $P_{light\_from\_cam}$ is the projection matrix to the light's reference frame from the camera's. We denote $(U^l_{cam}, V^l_{cam}, \hat{\textbf{Z}^l}_{cam})$ as the pixels and depth in camera's frame (subscript) projected into the light's frame, denoted by the superscript $l$. We index the shadow map, $\hat{\textbf{Z}}_{light} \in \mathbb{R}^{H\times W}$, at the projected camera pixels to retrieve the depth of the projected camera pixels from the light source. This is denoted as $\hat{\textbf{Z}}^{U^l_{c}, V^l_{c}}_{light}$ which is the shadow map indexed at pixel locations $U^l_{c}, V^l_{c}$. In practice, not all pixels will project within the shadow map's height and width constraints specified at the start of training. In graphics, these pixels are usually ignored, however, we clamp all our projections to lie within the height and width bounds to maintain differentiability. 

\noindent
\textbf{Soft Comparison.} Once we have the depths to the projected camera pixels and the depths from the light source to those pixels in the same reference frame, we can then compare them to discover if the camera pixel is in shadow. As illustrated by Figure \ref{fig:shadow_mapping}, if the depth from the light source to a point is less than the depth from the camera projected into the light's frame, it means that the light ray must have intersected an object before reaching that point. Thus, that point must be in shadow. Based on this logic, we formulate a soft comparison, which compares different depths to output the predicted binary shadow mask as follows:
\begin{equation}
\begin{gathered} 
\label{eq::soft_comp_1}
    \Delta \hat{Z}_{light} = \bigg ( \hat{\textbf{Z}^l}_{cam} - \hat{\textbf{Z}}^{U^l_{c}, V^l_{c}}_{light} \bigg) \\
    \hat{\textbf{M}}_{binary} = \textbf{max} \bigg ( \frac{\Delta \hat{Z}_{light}}{\beta}, \epsilon \bigg )
\end{gathered} 
\end{equation}
\noindent
We denote $\hat{\textbf{M}} \in \mathbb{R}^{H\times W}$ as the output of the entire pipeline: predicted shadow masks. The input to our soft comparison is the projected camera z-buffer into the light's frame, $\hat{\textbf{Z}^l}_{cam}$, and the shadow map indexed at the projected points $\hat{\textbf{Z}}^{U^l_{c}, V^l_{c}}_{light}$ from the \textbf{Projection} step. $\beta$ is a scaling hyper-parameter used to enlarge or decrease the difference, and $\epsilon$ is a threshold. We also formulate a ``smoother" version of the predicted shadows:
\begin{equation}
\begin{gathered} 
\label{eq::soft_comp_2}
    \hat{\textbf{M}}_{smooth} = \textbf{S} \big ( \textbf{normalize} \big ( \Delta \hat{Z}_{light}, \mu_{min}, \mu_{max} \big ) \bigg)
\end{gathered} 
\end{equation}

\noindent
Here, $ \mu_{min}, \mu_{max}$ are used to control the normalization function and \textbf{S} is the sigmoid function.

\subsection{Optimization}
\label{sec::optimization}
To enable convergence, we smooth the binary ground truth shadow masks $\textbf{M}$ to better guide the framework in predicting accurate shadow masks.

\noindent
\textbf{Distance Transform.} Binary images contain limited information for differentiation as the gradient is zero everywhere except for the edges where it is one. To encourage our model to estimate better shadow masks, thereby learning a better 3D model, we use a distance transform on the ground truth shadow masks. Specifically, we scale pixel intensities of a binary shadow mask by their distance to the nearest shadow edge. We modify the weighted distance transform in \cite{ronneberger2015unet} for our approach. The transformed binary shadow mask, $w(\textbf{M}, \sigma) = \textbf{M}_{w}$ is computed as follows: 
\begin{equation}
\begin{gathered} 
\label{eq::sigma_loss}
    w(\textbf{M}, \sigma) = \textbf{M} + \bigg( \emph{w}_c(\textbf{M}) + \emph{w}_{0} \cdot \text{exp}\big( -\frac{(d_1(\textbf{M}) + d_2(\textbf{M}))^2}{2\sigma^2} \big) \bigg)
\end{gathered} 
\end{equation}

Here, \textbf{M} is the ground truth binary shadow mask computed after applying a fixed threshold on binary images. $\emph{w}_c$ is weight map to balance class frequencies, $\emph{w}_0$ and $\sigma$ are hyper parameters. $d_1$ and $d_2$ are distances to the nearest and second nearest cell, respectively. We note from our experiments that this particular distance transform yields the most consistent convergence compared to other distance transforms, such as blurring. 

\noindent
\textbf{Shadow Mapping Loss.} We optimize our entire framework on binary shadow masks and train the MLP on the following loss:
\begin{equation}
\begin{gathered} 
\label{eq::sm_loss}
    \mathcal{L}_{sm} = ||w(\textbf{M}, \sigma) - \hat{\textbf{M}}||^2
\end{gathered} 
\end{equation}

Here, $w(\textbf{M}, \sigma)$ is the $\sigma$ weighted ground truth shadow mask, and $\hat{\textbf{M}}$ is the predicted shadow mask from Equation \eqref{eq::soft_comp_2}.

\begin{figure}[t]
    \includegraphics[width=0.85\linewidth]{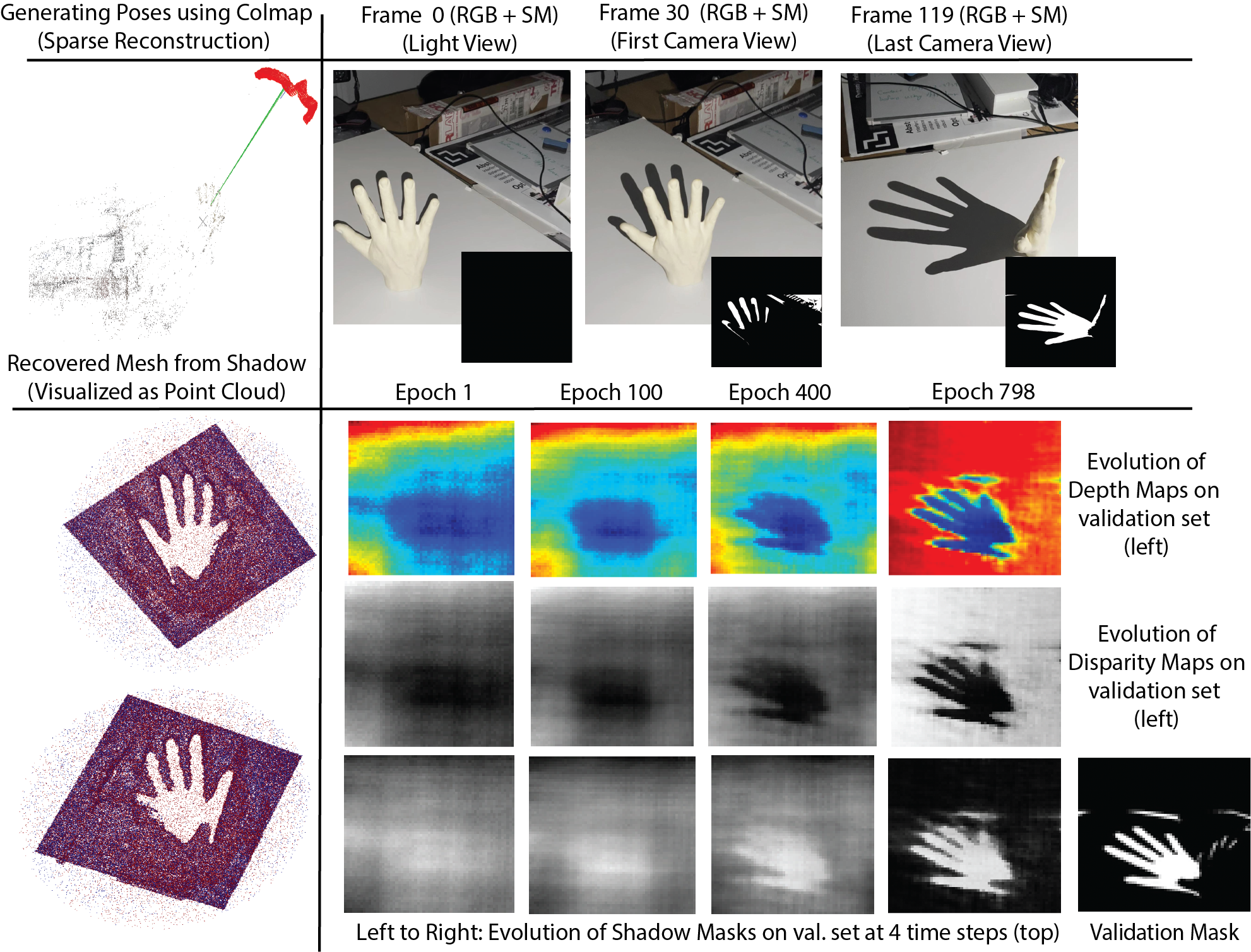}
    \centering
    \caption{\textbf{Real-World Experimentation:} We use the \emph{exact same pipeline and training scheme to reconstruct a 3D mesh from real-world data}. We take a video on the iPhone to generate poses for light and camera using COLMAP [43](\href{https://drive.google.com/file/d/1Bqi5WQqUoA062zyhrKAL9ntS0bnQhzga/view?usp=sharing}{video link}) and extract shadows using an intensity threshold. We show that our method can reconstruct a finer mesh of the hand from the real-world images. We highlight that our method can more easily generalize from sim2real in comparison to photometric approaches since we learn from only shadow masks, which are invariant to many real-world effects, such as texture.}
    \label{fig::real_world}
\end{figure}

\section{Implementation}
\label{sec:Implementation}
\subsection{Dataset}
We create a dataset of objects, including a cuboid, vase, chair and bunny in blender and render images of size $800\times800$. Although our approach does not have any constraints on the number or positions of light and camera, we fix one light source and randomly sample 200 camera positions on along the upper half of a sphere around the object. In simulation, we only consider top down/satellite views with the light source being very far away from the camera and the object to represent a distant ``sun-like'' source although this assumption was relaxed in real world dataset. Our dataset motivates the use of shadows for 3D reconstruction because in overhead imagery, vertical surfaces, like facades, are sampled poorly, whereas oblique lighting can expose this geometry. Link to the dataset is provided in the supplementary section. Details about the real world dataset is also provided in the supplementary section. 

\subsection{Training Details} 
We use a faster implementation of NeRF \cite{nerf} from \cite{queianchen_nerf}, which uses PyTorch Lightning as its backend \cite{NEURIPS2019_9015} \cite{falcon2019pytorch}. We down sample all images from $800\times800$ to $64\times64$ to fit on one RTX-3080 GPU. We use the same positional encoding scheme, $\gamma(x)$ and the MLP configuration $f_{\theta}$ used in \cite{nerf}. For camera projections, we write a custom \textit{Planar Projection Camera} class that encapsulates the projections and readily works with OpenGL and blender cameras. We gradually decreases the $\sigma$ (Equation \ref{eq::sigma_loss}) from $\{150, 100, 50\}$ during training to encourage the network to learn coarse to fine geometry. We also train a Vanilla NeRF model on RGB images of the same scene on resolution $64\times64$. To reconstruct the meshes, we run marching cubes on the learned implicit representations. More information on the exact training details is given in the supplementary section, including details about our more efficient differentiable shadow mapping implementation, which decreases the training time by half.

\section{Results}
\label{sec:Results}

\begin{table*}[t]
\centering
\begin{tabular}{p{0.15\linewidth}p{0.35\linewidth}p{0.35\linewidth}}
\hline
\textbf{Scene} & \textbf{RMSE Shadow Mesh} & \textbf{RMSE Vanilla NeRF}\\
\hline \hline
Cuboid & \textbf{0.0078} & 0.097 \\
Vase & \textbf{0.010} & 0.0.011 \\
Bunny & 0.0109 &  \textbf{0.0106}\\
Chair & \textbf{0.0092} & 0.0096\\
\hline
\end{tabular}
\caption{We quantitatively analyze the quality of the reconstructed meshes by running ICP \cite{icp1} on meshes generated by our proposed method, which only uses binary shadows masks, and meshes generated by a vanilla NeRF trained on full RGB images. We show RGB images from Vanilla NeRF in the supplementary along with training details.}
\label{table:icp_results}
\end{table*}


\noindent
\textbf{Evaluation Details.} We evaluate the performance of our method using root mean square error (RMSE) between the predicted point cloud and the ground truth point cloud, acquired with the iterative closest point (ICP) \cite{icp1} algorithm, as reported in Table \ref{table:icp_results}. In addition, we assess the visual quality of the predicted depth and shadow masks, and surface mesh, as shown in Figures \ref{fig::results_1} and \ref{fig::depth_fig}. The thresholds used to get the mesh from a volumetric representation are given in supplementary material. 

\begin{figure}[t]
\begin{center}
  \includegraphics[width=0.8\linewidth]{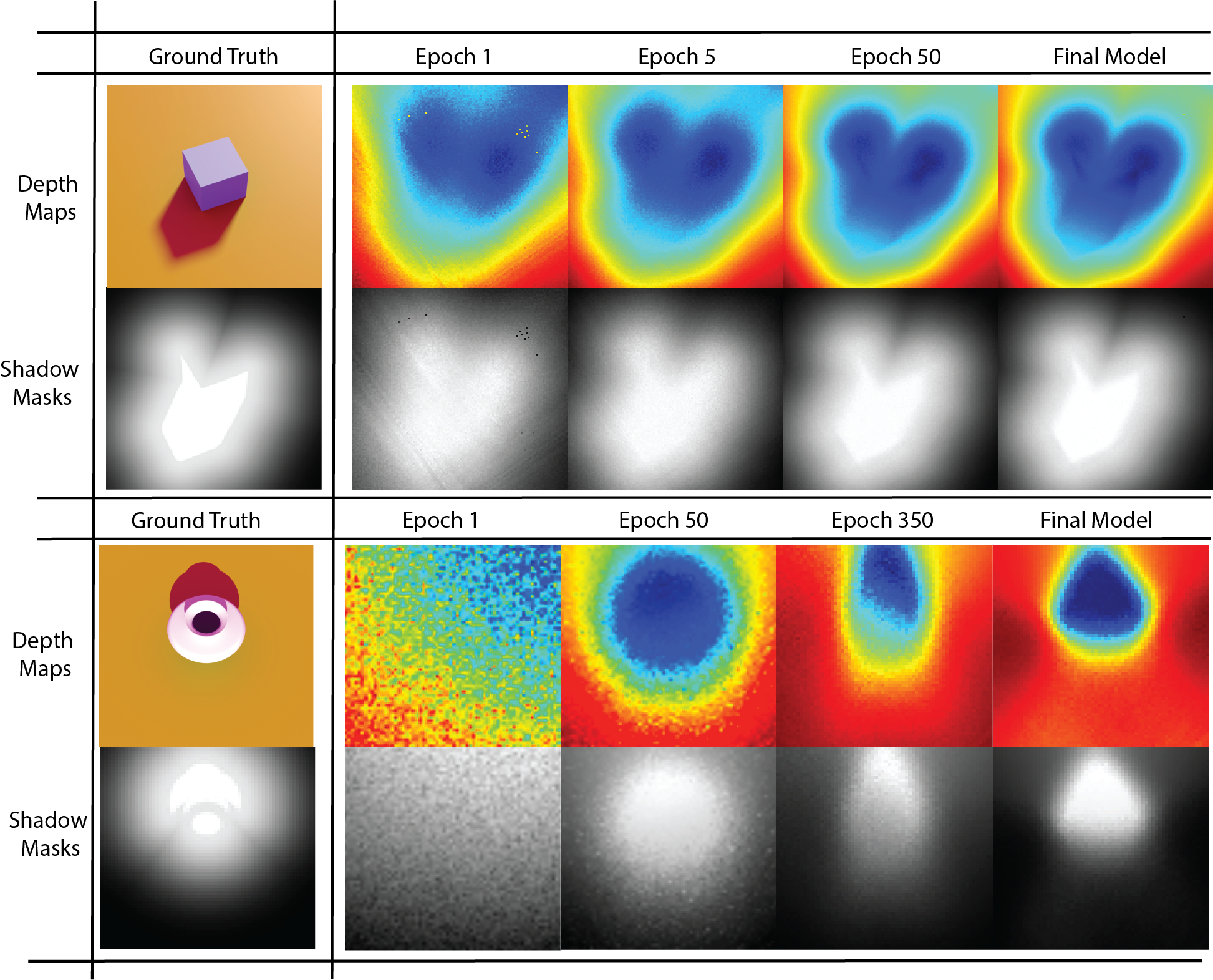}
\end{center}
\caption{\textbf{Evolution of depth maps during training through a novel camera viewpoint.} We visualize how the proposed forward model and differentiable rendering framework quickly converges to localize the object based on sparse shadow cues present and then slowly refines the coarse estimate on novel camera viewpoints. We believe that these results reveal that the differentiable volumetric rendering is a powerful framework that can rely on exploiting such physical cues to infer scene information. }
\label{fig::depth_fig}
\end{figure}

\noindent
\textbf{Simulated 3D Reconstruction Results.} We show the learned scene representations qualitatively by converting them to explicit meshes and rendering them using a signed distance function (SDF). Figure \ref{fig::results_1} shows the estimated meshes from our method on four object types. We compare our meshes to meshes generated by running vanilla NeRF on RGB images, and the ground truth by running marching cubes on the volume. Our datasets are rendered with overhead camera viewpoints, which enables shadows to be exploited. Given the binary and sparse nature of shadow masks in terms of their information content, we observe that our forward model coupled with the differentiable rendering framework converges to good coarse estimates of object geometry. Moreover, in the case of vases, the mesh reconstruction benefits from exploiting shadows as the algorithm can use \textit{hidden} cues present in the scene, such as the curvature of the vase, which are only partially visible when relying on photometric cues. We also show predicted depth maps and shadow masks on novel camera viewpoints not used during training in the supplementary materials.

\noindent
\textbf{Real-World Reconstruction Results.}
We show our method's ability to converge to a fine mesh on real-world data of a hand in Fig. \ref{fig::real_world}. Information on data acquisition is provided in the supplementary materials. We first note that our method is robust to coarse light poses as there are visible shadows from the estimated light's pose in Fig. \ref{fig::real_world} (please refer to the supplement for details). Our method is able to converge to a fine mesh of the hand, including the fingers and the space between them. We use only 74 shadow masks which makes our method versatile to environments with limited camera views and rarer objects, and no object priors. Moreover, we also show the convergence of the estimated shadow masks, disparity and depth maps from a novel viewpoint. The final estimated shadow mask shown in Fig. \ref{fig::real_world} is similar to the validation shadow mask and also contains some shadow artifacts due to the threshold segmentation. Additionally, the sim2real gap is not present for our pipleine as it only uses object shadows.

Lastly, we briefly discuss how the data-driven components finds the easiest solution that is consistent with our physics-driven forward model but not the actual world. We note that our training data only has views of cast shadows and does not contain any self-shadows which are present on the back of the hand. This causes the algorithm to instead estimate the mesh of the table and create a hollow imprint of the hand such that the specified shadow constraint is met. We note that a stand-alone mesh of the hand can be recovered from this imprint and that the recovered mesh is a possible solution given the shadow masks and proposed model. Imposing object priors or adding an extra view of self-shadow to the training set could result in a stand-alone mesh. 

\noindent
\textbf{Novel Viewpoint Rendering.} We observe predicted depth and shadow masks rendered from novel viewpoints in Figure \ref{fig::depth_fig}. The depth maps converge quite quickly to localize the object even when optimizing on the sparse physical cue of shadows. We posit that this convergence shows how powerful differentiable rendering is for exploiting physical cues to enable better 3D reconstruction. The depth maps converge slowly and we nudge the convergence by gradually decreasing the sigma values for the distance transform. The use of the distance transform leads to blurrier boundaries, however, the rendered mesh shows that a reasonably coarse 3D estimate is captured. 

\noindent
\textbf{Quantitative Analysis.} We also run our datasets on a vanilla NeRF \cite{nerf} implementation \cite{queianchen_nerf}. At lower resolutions and overhead viewpoints, we see that the NeRF approach fails to provide a reasonable fine mesh. We believe this failure is due to the down-sampling of images to $64 \times 64$, which may also be a reason as to why our meshes fail to capture fine details. We run ICP \cite{icp1} on the generated points cloud and show on-par results to the NeRF approach. Our goal, however, is not to outperform NeRF but to show the effectiveness of differentiable rendering framework in exploiting physical cues instead of ignoring them. The main takeaway from Table \ref{table:icp_results} is that differentiable volumetric renderers do not need to rely on 8 bit RGB information to reconstruct accurate meshes, but can also leverage other sources of information in the image in addition to relying on photometric cues. 

\noindent
\textbf{Limitations.} In cases such as the cuboid and the vase, we observe that the renderer converges to a predicted mesh that minimizes the shadow masks and the predicted shape even though it is typically a coarse estimate that envelopes the entirety of the object. This means that we see artifacts such as the pointed curve in the vase mesh, or the curvature of the bunny.  Since our algorithm only has geometry information where the binary shadow mask is true, areas that are never in shadow have no surface, which leads to incomplete meshes. Imposing a prior can be a solution to this problem. Moreover, our method also assumes known lighting position, which may not always be available. 

\section{Discussion}
\textbf{Exploiting Physical Cues.} One of the major goals of our work is to propose a framework within neural rendering that can readily exploit and learn from, instead of ignore, sparse physical cues, such as shadows. We believe that Fig. \ref{fig::real_world} shows that sparse physical cues like shadows, actually encode a lot of \textit{hidden} information about the scene and can indeed be exploited. By constructing explicit differentiable forward models and leveraging gradient-friendly volumetric rendering, we can exploit these cues in conjunction with relying on photometric consistency between images. 

\noindent
\textbf{Differentiable Shadow Rendering.} In rasterization, shadow computation is done through shadow mapping as it is well suited and efficient. However, shadow computation in ray tracing are expensive as every ray needs to compute a path to the light source. Therefore, many ray tracing approaches also use shadow mapping to compute shadows efficiently. We use shadow mapping in our neural rendering approach as well. Our approach is similar to the shadow mapping in graphics as it assumes a binary label on shadows and does not consider soft shadows or ambient lighting. However, we invert shadow mapping and exploit it to do 3D reconstruction, not to render photorealistic images. Moreover, our approach is readily extendable to varying light and camera sources. 


\subsection{Future Work}
We observe that volumetric rendering can converge onto coarse estimates of the object geometry by only relying on shadows, and can be extended to problems such as non-line-of-sight imaging (NLOS) \cite{Velten12recoveringthreedimensional} and imaging behind occluders \cite{ibo-sfs}. As shadows themselves are never the only cue present to reconstruct the scene, our work can also be easily integrated with existing NeRF approaches that rely only on photometric cues as our shadow loss \ref{eq::soft_comp_2} can be used as a regularizer or an  auxiliary loss, especially as shadows are invariant to viewpoint changes, surface reflectance properties, or texture changes. 

\subsection{Conclusions}
We show that modern neural rendering techniques can learn neural scene representations (neural shadow fields) and encode 3D geometry just from binary shadow masks. We are motivated by traditional shape-from-X algorithms that typically construct physics-driven inverse models that can exploit cues for 3D reconstruction. We observe that data-driven neural rendering frameworks ignore cues such as shadows, relying on on photometric cues instead. We thus propose a graphics-inspired differentiable shadow rendering component that leverages a  volumetric renderer to encode a scene solely from its shadows. 

\noindent 
\textbf{Acknowledgements.} This research was supported by the SMART Contract IARPA Grant \#2021-20111000004. We would also like to thank Systems \& Technology Research (STR). In addition, the authors would also like to thank Professor Voicu Popescu (Purdue University) for being so generous with his time and the valuable discussions that came from our meetings. 

\clearpage
%
%
\bibliographystyle{splncs04}
\bibliography{egbib}
\end{document}


\pagestyle{headings}
\mainmatter
\author{}
\authorrunning{Tiwary K. et al.}
%
\institute{}



\renewcommand{\thefigure}{A.\arabic{figure}}
\renewcommand{\theequation}{A.\arabic{equation}}
\renewcommand{\thesection}{A.\arabic{section}}


\section{Additional Details}

Reconstructing any 3D representation just from the object's binary shadow masks is a very hard and unsolved problem. Shadows possess important geometric information about the object, however, due to the ill-posedness, many different kinds of objects that have similar appearance in 2D can cast the same shadow. Moreover, these objects can map to the same shadow mask and have very different underlying 3D shapes making this problem more difficult than recovering 3D geometry from images. Our algorithm makes limited assumptions about the underlying scene working with \emp{hard shadows}. It also doesn't assume the number of objects in the scene and tries to fit the best 3D model that explains the shadows masks. 

\begin{figure}
\begin{center}
  \includegraphics[width=0.9\linewidth]{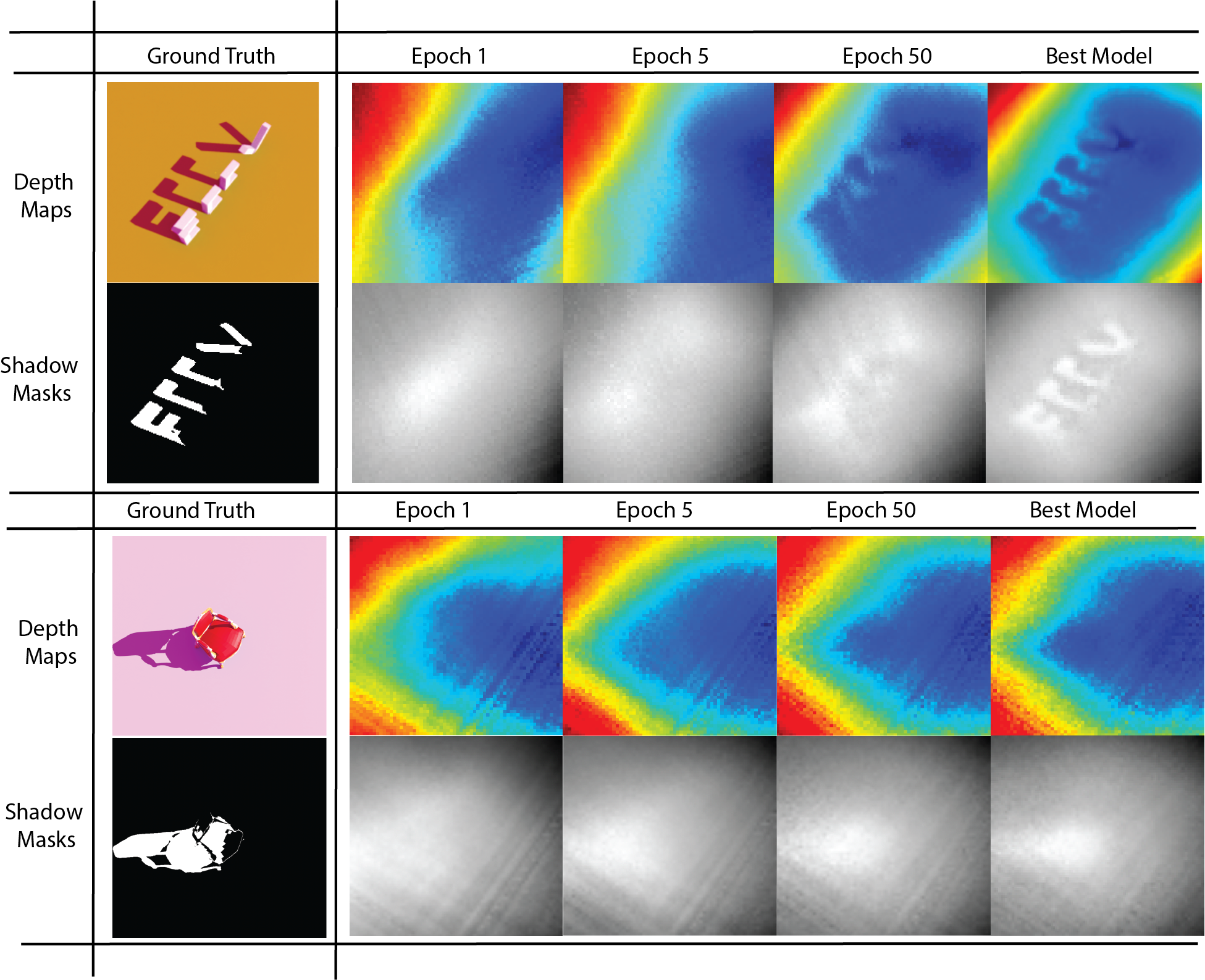}
\end{center}
\caption{\textbf{Predicted depth maps and shadow masks for a scene with multiple objects and a chair during training on validation poses.} Reconstructing any 3D representation from the object's binary shadow masks is a very hard and we show that we can extend our algorithms to a scene with \textbf{arbitrary} number of objects. This is due to our use of implicit volumetric representation and that our our algorithm makes limited assumptions about the scene or its objects.}
\label{fig::depth_figs}
\end{figure}

\section{Results}
We show predicted depth and shadow masks on a scene with multiple objects and the chair in Figure \ref{fig::depth_figs}, and on a bunny object in Figure \ref{fig::suppl_bunny}. In the first row of Figure \ref{fig::depth_figs}, we run our method on a scene with multiple objects that spell \textbf{ECCV}. If we had used explicit meshes, we would have to specify the number of objects before training, however, by parametrizing the scene with opacities, we can scale up to an arbitrary number of objects placed in any position in the scene. For the scene with multiple objects, we generate shadow masks and feed them to the proposed algorithm as input, without specifying additional information. Since our algorithm does not assume any prior knowledge about the scene or its objects, it can optimize over and find the best volumetric densities that explain the shadow mask. This result further shows how traditional algorithms like shape-from-shadows/X can benefit from differentiable volumetric rendering as with it they can scale to an arbitrary number of objects and more complex scenes. 

For the chair object in Figure \ref{fig::depth_figs}, observe how the chair is on the right of the casted shadows and the model accurately predicts and segments out that area, and is able to infer the sharp corner. 

As shown in Figure \ref{fig::suppl_bunny}, the model is able to learn the coarse shape and localize the bunny well through its shadow map. However, it fails to carve our the ears and the and finer details. Our shape-from-shadow algorithm has the poorest performance on the bunny (Figure \ref{fig::suppl_bunny}), most likely due to its complexity, curved surfaces, and its protruding ears that can imply many different possible combinations of the underlying 3D shape.  Note that the model is trained on coarser versions of the fine shadow masks (shown in first column), as described by our training scheme in Supplementary Section \ref{sec::sup_impl}.


\begin{figure}[t]
\begin{center}
  \includegraphics[width=0.5\linewidth]{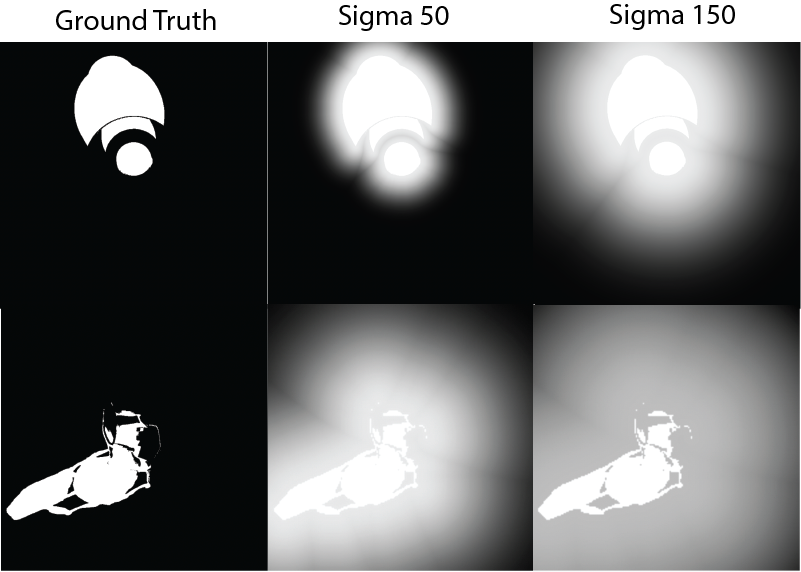}
\end{center}
\caption{\textbf{Increasing the distance transform weight "sigma" hyperparameter.} We show the shadow maps that the model actually learns from. We apply the distance transform and slowly decrease it in our training. The transform helps with taking smoother gradients w.r.t. the model's parameters.}
\label{fig::sigma_thres}
\end{figure}

\section{Implementation}
\label{sec::sup_impl}
\begin{figure}[t]
\begin{center}
  \includegraphics[width=0.9\linewidth]{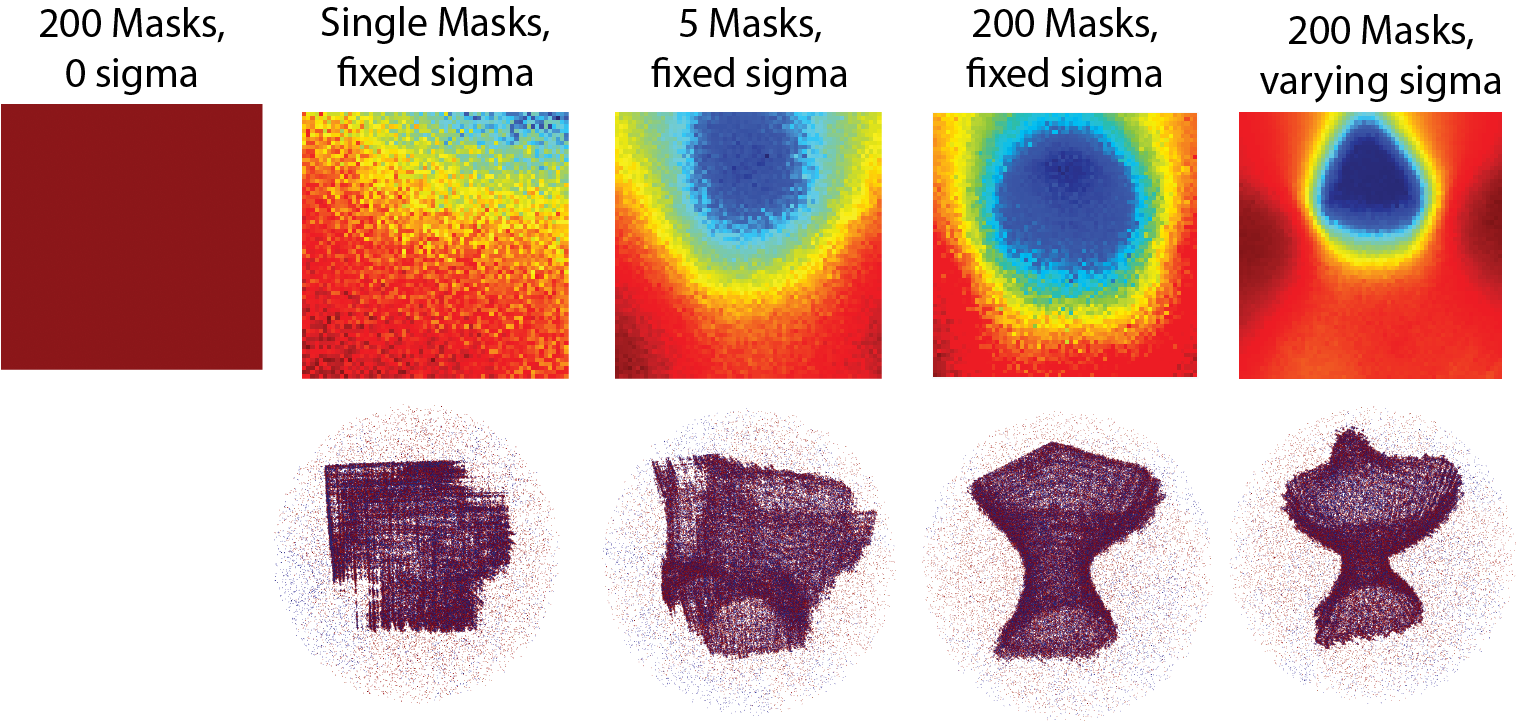}
\end{center}
\caption{\textbf{Ablation on sigma and number of shadow masks.} We study the importance of the distance transform in our method. Without the distance transform to guide and smoothen the shadow masks, there is no convergence in the depth maps or the mesh. We notice an iterative coarse to fine mesh estimation done by the framework as we increase the number of masks and vary sigma. We observe that the number of masks carves the space down to create finer meshes. Varying sigma with many shadow masks yields the best results, although it can overfit to the shadow mask and cause mesh artifacts.}
\label{fig::ablation_case}
\end{figure}

\textbf{Model Details.} During training, we set $\beta$ $1\textbf{e}^{-2}$ or $1\textbf{e}^{-3}$ and $\epsilon$ is usually set to $0$. We set the $ \mu_{min}, \mu_{max}$ to be 0 and 1 respectively and compute gradients directly on the normalized values instead of using the sigmoid. For the position encoding we also use parameters from the original NeRF implementation. We use an 8-layer MLP to parametrize our scene. To enable better depth reconstruction we also use a coarse and a fine MLP, the coarse is sampled 64 times and the fine 128. Our MLP, however, does not have an extra head for the color and does not take the viewing direction as input. 

\textbf{Training Scheme.} To train the model, we use the PyTorch implementation of the Adam optimizer \cite{adam_opt}, use a learning rate of $5\times 10^{-4}$, and use a step function to decay the learning rate at 20 epochs. We evaluate our model on the validation data, and test it against real meshes from blender. We start training with a high distance setting for our distance transform $\sigma = \{100,150\}$ and train for 150 epochs. Then we lower the $\sigma = \{50, 45\}$ to continue training. We note that using a $\sigma$ value lower than this causes our model to diverge and start to learn spurious depth maps. Our entire implementation trains on one Tesla V-100 and takes a half a day. The models were typically trained to run for 200/300 epochs where each epoch runs on all rays generated from all camera pixels. Only the cuboid was trained on 2 Tesla V100s on an image of size $128\times128$ with 64 fine and coarse samples. This training took 3 days to complete. 

\textbf{Real-World Dataset Pipeline.} We place a mannequin object of the hand in the scene and use the flashlight from a smartphone and take a video of the scene using another smartphone. We start the data collection near the flashlight so that the first 30 poses are roughly similar to the pose of the light. We process the remaining 74 frames and perform an intensity threshold to extract the shadow masks. This simple technique also causes the darker regions to be classified as shadows, as visible in Fig. \ref{fig::real_world}. We then run COLMAP \cite{schoenberger2016sfm} \cite{schoenberger2016mvs} on the video to extract the poses, and use the first frame's pose as the light pose discarding the rest. We note that our method is robust to the coarsely estimated light pose as there are shadows visible from that viewpoint. 

\textbf{Efficient Differentiable Shadow Rendering.} Our approach requires two NeRF forward passes per epoch to train: one from the light's perspective and one from the camera's. To make training faster, we implemented a more efficient method to compute depths. We exploited the fact that our method requires the shadow map array to be indexed by the projected camera pixels, therefore we only need to compute a full shadow map on $H\times W$ rays and can batch the camera rays. This approach worked well and decreased our computational cost by roughly half. Moreover, our method does not make any assumption on the size of the camera and light depth arrays, and therefore we could also use a smaller shadow map. We also implemented a more efficient method of projecting the camera pixels into the light frame of reference, only computing the light depths on the projected locations instead of computing the full shadow map. This implementation, however, does not lead to any convergence, indicating that the loss computed on out-of-shadow pixels is also important and gives valuable information that helps in carving away that space to refine the mesh. 

We also experimented with changing the number of fine samples used to sample opacities for a given light ray, computing gradients on the light opacities every K steps, in addition to sampling light depths at varying intervals instead of every step. Albeit many such methods did speed up training, we decided to use a basic setup which samples the light and camera rays with 64 coarse and 128 fine samples every iteration, with gradients being computed at each step.

\begin{figure}[t]
\begin{center}
  \includegraphics[width=0.8\linewidth]{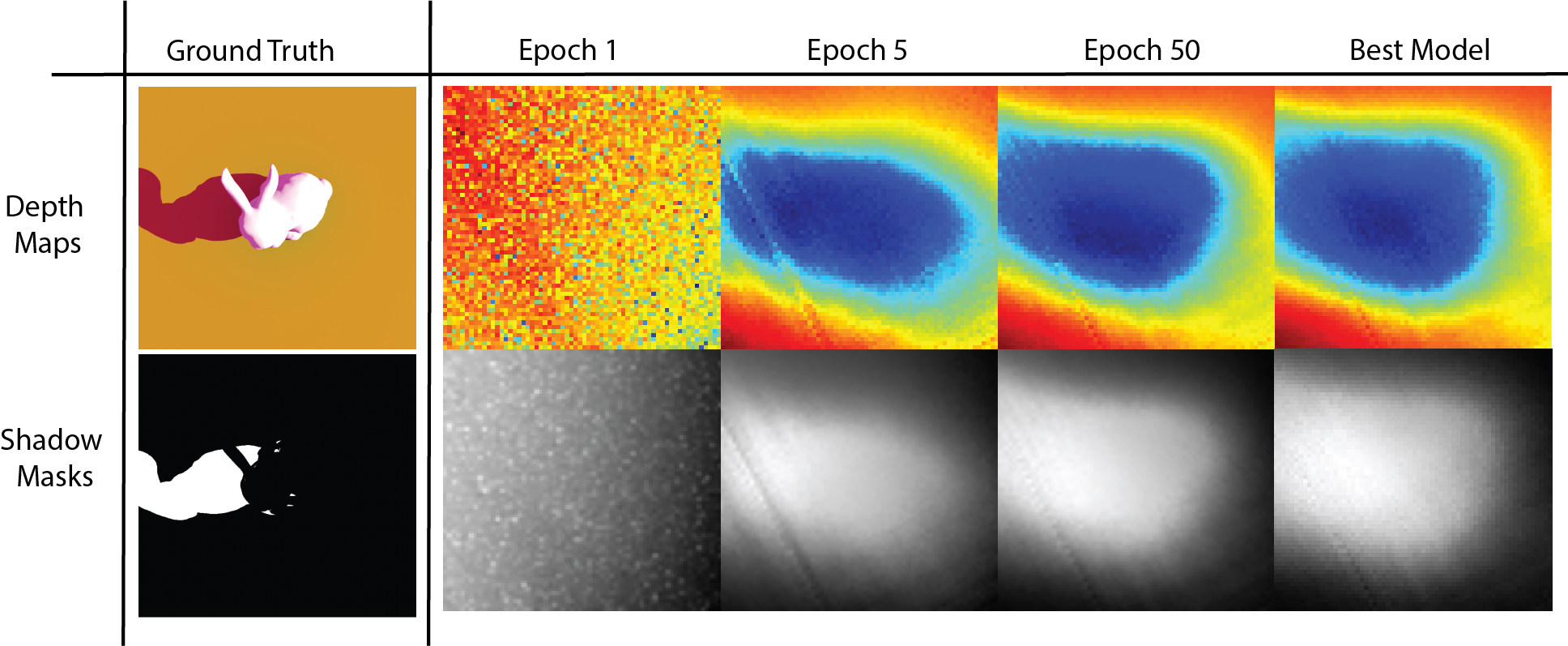}
\end{center}
\caption{\textbf{Predicted depth maps and shadows masks during training on the bunny's validation poses.} Even though the model is able to learn the greater shape and localize the bunny well through its shadow map, its poor performance is probably due to its complexity, curves surfaces and its protruding ears that can imply many different possible combinations of the underlying 3D shape.}
\label{fig::suppl_bunny}
\end{figure}

\textbf{Explicit Mesh from Implicit Representations.} Extracting an explicit mesh from an implicit volumetric representation typically involves running marching cubes \cite{lorensen1987marching}. We used the PyMCubes library implementation of the marching cubes algorithm. To do this, we create a bounding volume around the object and query the 3D volume for opacities at points inside the cuboid. In practice, we query along each dimension 128 to 256 of the volume of size $(H,W,D)$ therefore creating voxels of size $H/128, W/128, D/128$. We query for an opacity at every voxel once to create a volume that can then be fed into the marching cubes algorithm. We use a set contour value of 0 to search for isosurfaces in the volume. This gives us a non-smooth, explicit mesh with vertices and triangular faces which is then used for visualization and evaluation. Note that the volume bounds are different for each object and we find them using trial and error. In $(x,y,z)$ direction, they are $\{\pm5\}$ for cuboid, $\{\pm35\}$ for chair, $\{\pm45\}$ for vase and $\{\pm35\}$ bunny. We refer our readers to our code for more information. Additionally, we do want to note that extracting explicit meshes from implicit representations is an extremely difficult task which requires a lot of trial and error. Even when results look great for novel view synthesis with the RGB NeRF framework, marching cubes does not produce perfectly smooth meshes. Often these meshes have jagged edges and protruding elements. Moreover, this becomes a tremendous problem when there is more than one object, as we start to extract extremely noisy meshes. However, alignment of explicit meshes is one of the most accurate metrics for 3D reconstruction evaluation, which is why we use it as an evaluation metric. 



\section{Ablation Studies}

We show in Figure 6 in the main text the final reconstructions from different experiments varying different parameters. We observe that starting from a smoothed-out shadow mask is critical for the algorithm to construct a coarse mesh and the algorithm almost never converges without smoothing applied to the initial binary shadow masks. This makes sense as the gradients are now non-zero around the edges of the binary mask, and can help guide the network in constructing a mesh that is consistent with all the shadow images. Smoothing techniques have also been used in \cite{NMR} and \cite{liu2019soft} where the goal was to recover 3D mesh from a single silhouette image. 
Since the light is fixed in the scene and unchanged, the amount of information available about the object's geometry is also fixed. The varying camera positions provide different viewpoints to the object and help the algorithm differentiably carve away that space. We note that even with five shadow masks our algorithm is able to reconstruct a coarse object, however it is not able to carve away the right and the left parts of the vase, which we believe is due to the lack of viewpoints from those poses. We note that 200 views of the object means that some positions are sampled more than once, and is more than enough to help the algorithm carve out the unnecessary elements in the mesh. We also show that varying the distance transform weight from 150 to 50, as shown in Figure 6 for the vase, can create crisper depth maps and faster convergence. However, this also leads to overfitting to the shadow mask since exactly fitting a mesh to the shadow does not necessarily translate to inferring the actual underlying shape of the object due to fundamental ill-posedness of the problem.

\begin{figure}[t]
\begin{center}
  \includegraphics[width=0.8\linewidth]{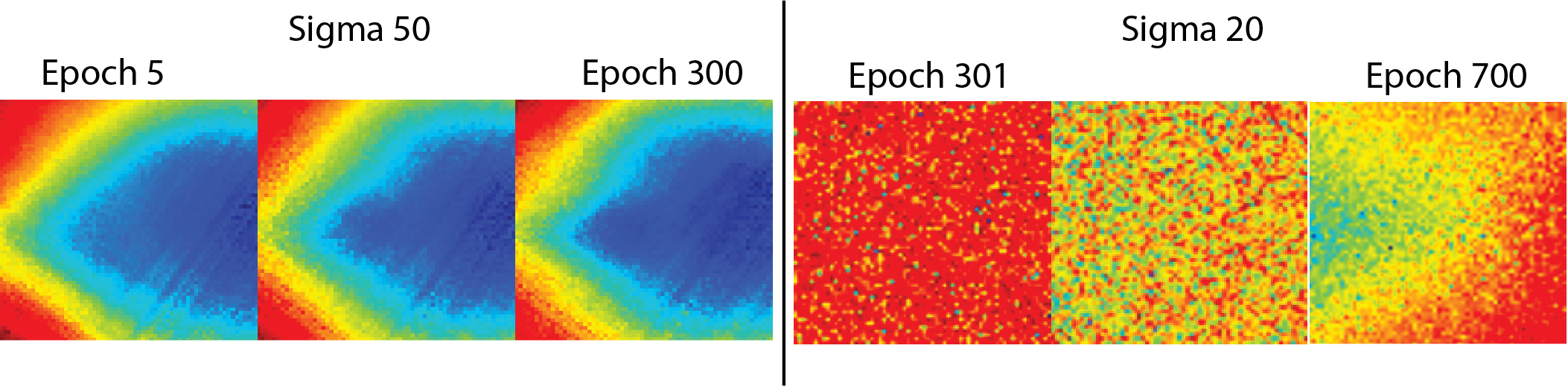}
\end{center}
\caption{\textbf{Drastically changing distance transform weight during training for the Chair Object.} This figure shows the challenge in using shadows with a gradient based technique. Even though we reduce the weight during training, drastically reducing it can cause the model to unlearn the scene representation as the information it relied on converge to the object is now missing.}
\label{fig::changing_sigma}
\end{figure}

In Figure \ref{fig::changing_sigma} we also show drastically reducing the distance transform weight from 50 to 20 can cause the model to unlearn the scene representation. This shows how little gradient information shadows contain and that the model needs to be consistently nudged in order for it to be guided to reconstruct the underlying mesh. 
